\documentclass{article}

\usepackage[final]{nips_2016}
\usepackage[utf8]{inputenc} 
\usepackage[T1]{fontenc}    
\usepackage{hyperref}       
\usepackage{url}            
\usepackage{booktabs}       
\usepackage{amsfonts}       
\usepackage{microtype}      
\usepackage{latexsym}
\usepackage{graphicx}
\usepackage{algorithm}
\usepackage{algorithmic}
\usepackage{amsmath}
\usepackage[normalem]{ulem}
\usepackage{color}
\usepackage{caption,booktabs,array}\usepackage{multirow}
\usepackage{subfig}
\usepackage{mwe}
\usepackage{setspace}

\usepackage{times}

\title{A Shallow High-Order Parametric Approach to Data Visualization and Compression}

\author{
Martin Renqiang Min$^{\ddagger}$, Hongyu Guo$^{\dagger}$, Dongjin Song$^{\ddagger}$ \\
$^{\ddagger}$ NEC Labs America, Princeton, NJ 08540\\
\texttt{renqiang@nec-labs.com, dosong@ucsd.edu}\\
$^{\dagger}$ National Research Council Canada, Ottawa, ON\\
\texttt{hongyu.guo@nrc-cnrc.gc.ca}\\
}

\begin{document}

\maketitle
\begin{abstract}

Explicit high-order feature interactions efficiently capture essential structural knowledge about the data of interest and have been used for constructing generative models. We present a supervised discriminative High-Order Parametric Embedding (HOPE) approach to data visualization and compression. Compared to deep embedding models with complicated deep architectures, HOPE generates more effective high-order feature mapping through an embarrassingly simple shallow model.  Furthermore, two approaches to generating a small number of exemplars conveying high-order interactions to represent large-scale data sets are proposed. These exemplars in combination with the feature mapping learned by HOPE effectively capture essential data variations. Moreover, through HOPE, these exemplars are employed to increase the computational efficiency of kNN classification for fast information retrieval by thousands of times. For classification in two-dimensional embedding space on MNIST and USPS datasets, our shallow method HOPE with simple Sigmoid transformations significantly outperforms state-of-the-art supervised deep embedding models based on deep neural networks, and even achieved historically low test error rate of 0.65\% in two-dimensional space on MNIST, which demonstrates the representational efficiency and power of supervised shallow models with high-order feature interactions.
\end{abstract}

\vspace{-2mm}
\section{Introduction}
\vspace{-2mm}
High-order feature interactions naturally exist in many real-world data, including images, documents, financial time series, biological sequences, and medical records, amongst many others. These interplays often convey essential information about the latent structures of the datasets of interest. For data embedding and visualization, therefore, it is crucial to utilize these high-order characteristic features to generate the dimensionality reduction function. 

Recently supervised deep learning models have made promising progresses on sensory data with a lot of regularities such as images and speeches, in terms of generating powerful complex parametric embedding functions that capture high-order feature interactions through deep architectures. Current state-of-the-art deep strategies, however, fail to deploy  an \textit{explicit high-order parametric} form to map high-dimensional data to low-dimensional space. Explicit parametric mapping not only effectively avoids the need to develop out-of-sample extension as in the cases of non-parametric methods such as the t-SNE~\cite{van2008visualizing},  but also reveals the structural information intuitively understandable to human that enables people to make good sense of the data through visualization or to acquire interpretative knowledge out of the visualization. 
Furthermore, current embedding methods often ignore or fail to perform data compression or summarization while generating embedding. Such functionality is very desirable when dealing with large-scale datasets for fast information retrieval, on which we often perform kNN classification and computational efficiency is important.

To address the above mentioned challenges, in this paper, we present a High-Order Parametric Embedding (HOPE) approach. The aims of HOPE are two-fold: learning an explicit high-order parametric embedding function for data visualization and constructing a small set of synthetic exemplars with high-order feature interactions borne to represent the whole input dataset. In specific, our approach targets supervised data visualization with two new procedures. Firstly, we linearly map explicit high-order such as $k$-order interaction features, which are the products of all possible $k$ features, to two-dimensional space for visualization, such that all pairwise data points in the same class stay together and pairwise data points from different classes stay farther apart. To avoid directly enumerating all possible $k$-feature interactions that is computationally prohibitive, we propose using tensor factorization to learn a set of feature interaction filters. As a result, the high-order interactions can not only be preserved in the low-dimensional embedding space, but also be explicitly represented by these feature interaction filters. Consequently, one can directly compute the explicit high-order interactions hidden in the data. Secondly, we develop exemplar learning techniques 
to create a small set of exemplars associated with the embedding to represent the entire data set. As a result, one can just use these exemplars to perform fast information retrieval such as the widely adopted kNN classification, instead of the whole data set, to speed up computation and gain insight into the characteristic features of the data. This is particularly important when the data set is massive. 
  
We evaluated the performance of HOPE and its nonlinear extension using the benchmarking MNIST and USPS datasets. Our experimental results strongly support the effectiveness and the efficiency of our methods for both data visualization and data compression. 

\vspace{-2mm}
\section{Related work}\label{sec:related}
\vspace{-2mm}

Dimensionality reduction and data visualization methods mainly fall into two categories, unsupervised approaches ~\cite{Belkin2003,Hinton2003,Hotelling1933,MinThesis,Roweis2000,Tenenbaum2000,Shieh2011,Maaten09,van2008visualizing} and supervised approaches~\cite{MCML2006,NCA2005,He2003,MinMYBZ10,MinICDM,Weinberger:2009}. Among the supervised approaches, MCML~\cite{MCML2006}, NCA~\cite{NCA2005}, LPP~\cite{He2003}, and LMNN~\cite{Weinberger:2009} are linear methods, and dt-MCML~\cite{MinMYBZ10}, dt-NCA~\cite{MinMYBZ10}, and DNN-kNN~\cite{MinICDM} are deep nonlinear methods. 
Our method HOPE is a supervised embedding approach. But unlike the above methods, it directly maps explicit high-order interaction features, instead of the original input features, either through linear projection or sigmoid transformations followed by linear projection to low-dimensional embedding space. This simple feature mapping enables users to identify important interaction features. With only linear projection, HOPE can be viewed as a linear method applied to high-order interaction features, so its baseline counterparts should be those linear embedding methods. With Sigmoid transformation followed by linear projection, it can be viewed as a shallow nonlinear method. 
Through HOPE, a small number of exemplars conveying high-order feature interactions are synthesized. 
It is worth noting that, HOPE with the proposed two exemplar learning techniques is similar to but intrinsically different from Stochastic Neighbor Compression (SNC)~\cite{SNC2014}. Specifically, learning exemplars in HOPE aims for constructing an embedding mapping that optimizes an objective function of maximally collapsing classes~\cite{MCML2006} instead of neighbourhood component analysis~\cite{NCA2005} in SNC. In particular, unlike in SNC, the joint exemplar learning technique in HOPE is coupled with high-order embedding parameter learning, which powers the exemplars created to capture essential data variations bearing high-order interactions. 
In addition, the results of HOPE with the k-means based exemplar learning technique show that, using a powerful feature mapping generated by HOPE with Sigmoid transformations, optimization over exemplars is unnecessary, which is actually against the motivations of SNC. 

High-order feature interactions have been studied for building more powerful generative models such as Boltzmann Machines and autoencoders~\cite{DBLP:journals/corr/GuoZM15,DBLP:conf/iccv/Memisevic11,DBLP:conf/aistats/MinNCG14,DBLP:conf/cvpr/RanzatoH10,DBLP:journals/jmlr/RanzatoKH10}. Factorization Machine (FM)~\cite{Rendle2010} and FHIM~\cite{Min2014kdd} are similar to the version of HOPE with only linear projection, but they used feature interactions for classification, regression, or feature selection. None of previous research has been conducted under the context of data embedding, visualization, or compression, and therefore has different objective function or parametric form. Especially, our joint learning approach is completely different from previous methods. And to the best of our knowledge, our work here is the first successful one to model input feature interactions with order higher than two for practical supervised embedding. 
\section{High-Order Parametric Embedding}\label{sec:method}
\vspace{-2mm}
\subsection{Supervised high-order parametric embedding by maximally collapsing classes}\label{sec:sup}
Given a set of data points $\mathcal{D} = \{{\mathbf x}^{(i)}, L^{(i)}: i = 1,\ldots, n\}$, where ${\mathbf x}^{i} \in R^H$ is the input feature vector with the last component being $1$ for absorbing bias terms, $L^{(i)} \in \{1, \ldots, c\}$ is the class label of labeled data points, and $c$ is the total number of classes. HOPE intends to find a high-order parametric embedding function ${\mathbf f}({\mathbf x}^{(i)})$ that maps high-dimensional data points $x_{i}$ to a low-dimensional space $R^h$ $(h < H)$, where we expect that data points in the same class stay tightly close to each other and data points from different classes stay farther apart from each other. For data visualization, we often set $h = 2$. Unlike previous methods that directly embed original input features ${\mathbf x}$, HOPE assumes that high-order feature interactions are essential for capturing structural knowledge and learns a similarity metric directly based on these feature interactions. Suppose that HOPE directly embeds $O$-order feature interactions, i.e., the products of all possible $O$ features $\{x_{i_1} \ldots x_{i_t} \ldots x_{i_O}\}$ where $t \in \{1, \ldots, O\}$, and $\{i_1, \ldots, i_t, \ldots, i_O\}\in \{1, \ldots, H\}$. A straightforward approach is to explicitly calculate all these $O$-order feature interactions and use them as new input feature vectors of data points, and then learn a linear projection matrix $\mathbf{U}$ to map them to a $h$-dimensional space as follows,
\begin{align}
\mathbf{y} = \mathbf{U}^T \left[ \begin{matrix}
x_1 \ldots x_1 \ldots x_1 \\
\vdots   \\
x_{i_1} \ldots x_{i_t} \ldots x_{i_O} \\
\vdots \\
x_H \ldots x_H \ldots x_H \\
\end{matrix} 
\right],
\end{align} 
where ${\mathbf U} \in {\mathbf R}^{H^O \times h}$, and $\mathbf{y} \in R^h$ is the low-dimensional embedding vector. We can rewrite the above equation in the following equivalent tensor form, 
\begin{equation}
\label{tensor}
y_s = \sum_{i_1 \ldots i_t \ldots i_O}\mathcal{T}_{i_1 \ldots i_t \ldots i_Os}x_{i_1} \ldots x_{i_t} \ldots x_{i_O},
\end{equation}
where $\mathbf{\mathcal{T}}$ is a $(O+1)$-way tensor, $s = 1, \ldots, h$. However, it's very expensive to enumerate all possible $O$-order feature interactions. For example, if $H = 1000, O = 3$, we must deal with a $10^9$-dimensional vector of high-order features. To speed up computation, we factorize the tensor $\mathbf{\mathcal{T}}$ as follows,
\begin{equation}
\mathcal{T}_{i_1 \ldots i_t \ldots i_Os} = \sum_{f=1}^F C^{(1)}_{fi_1}\ldots C^{(t)}_{fi_t}\ldots C^{(O)}_{fi_O} P_{fs},  
\end{equation}
where $F$ is the number of factors. If we enforce ${\mathbf C}^{(1)} = \ldots = {\mathbf C}^{(t)} = \ldots = {\mathbf C}^{(O)} = {\mathbf C}$, the $s$-th high-order embedding coordinate in Equation~\ref{tensor} can be rewritten as follows,
\begin{eqnarray}
\label{factensor}
\scriptsize
y_s & = & \sum_{i_1 \ldots i_t \ldots i_O} \sum_{f=1}^F C^{(1)}_{fi_1}\ldots C^{(t)}_{fi_t}\ldots C^{(O)}_{fi_O} P_{fs} x_{i_1} \ldots x_{i_t} \ldots x_{i_O}, \nonumber \\
& = & \sum_{f=1}^F P_{fs}(\sum_{i_1=1}^H C^{(1)}_{fi_1} x_{i_1}) \ldots (\sum_{i_t=1}^H C^{(t)}_{fi_t} x_{i_t}) \ldots (\sum_{i_O=1}^H C^{(O)}_{fi_O} x_{i_O}) \nonumber \\
& = & \sum_{f=1}^F P_{fs}(\sum_{i=1}^H C_{fi} x_{i})^O = \sum_{f=1}^F P_{fs}({\mathbf C_f}^T {\mathbf x})^O,
\end{eqnarray}
where $s = 1, \ldots, h$. With the above constrained tensor factorization, we can easily calculate the linear embedding for any high-order interaction features of any high-dimensional data by an embarrassingly simple operation, that is, a linear projection followed by a power operation. 
It is worth noting that, the above factorization form not only reduces computational complexity significantly, but also is amenable to explicitly model different order of feature interactions in the data with a user-specified parameter $O$. 

The above HOPE method has an explicit high-order parametric form for mapping and is essentially equivalent to a linear model with all explicit high-order feature interactions expanded as shown above. Compared to supervised deep embedding methods with complicated deep architectures, the above linear projection method has limited modeling power. Fortunately, there is a very simple way to significantly enhance the model's expressive power. That is, by simply adding Sigmoid transformations to the above factorized model before performing linear projection. We call the resulting model Sigmoid HOPE (S-HOPE). In S-HOPE, the $s$-th coordinate of the low-dimensional embedding vector $\mathbf y$ is computed as,
\begin{equation}
\label{shopemap}
y_s = \sum_{k=1}^m P_{sk} \sigma(\sum_{f=1}^F w_{fk}({\mathbf C_f}^T {\mathbf x})^O + b_k),
\end{equation}
where $b_k$ is the bias term and $\sigma (x) = \frac{1}{1 + e^{-x}}$. S-HOPE dramatically improves the modeling power of HOPE with a trivial modification. As is shown in the experimental result section, the resulting shallow high-order parametric method even significantly outperforms the state-of-the-art deep learning models with many layers for supervised embedding, which clearly demonstrates the representational power of shallow models with high-order feature interactions. 

Given the high-order feature mapping ${\mathbf y}^{(i)}$ of the $i$-th data point ${\mathbf x}^{(i)}$, $i = 1, \ldots, n$, we perform supervised metric learning by maximally collapsing classes (MCML)~\cite{MCML2006}. Following the line of research in~\cite{MCML2006,NCA2005,Hinton2003,   MinMYBZ10}, we deploy a stochastic neighbourhood criterion to compute the pairwise similarity of data points in the transformed space.  In this setting, the similarity of two data points ${\mathbf y}^{(i)}$  and ${\mathbf y}^{(j)}$  are measured by a probability $q_{i|j}$. The $q_{j|i}$ indicates the chance of the data point ${\mathbf y}^{(i)}$  assigns ${\mathbf y}^{(j)}$  as its nearest neighbor in the low-dimensional embedding space. Following the work in~\cite{MinMYBZ10}, we use a heavy-tailed t-distribution to compute $q_{j|i}$ for supervised embedding due to its capabilities of reducing overfitting, creating tight clusters, increasing class separation, and easing gradient optimization. Formally,  this stochastic neighborhood metric first centers a t-distribution over ${\mathbf y}^{(i)}$, and then computes the density of ${\mathbf y}^{(j)}$ under the distribution as follows.
\begin{eqnarray}
q_{j|i} & = & \frac{(1 + d_{ij})^{-1}} {\sum_{kl:k \neq l}  (1 + d_{kl})^{-1}}, \quad q_{ii} = 0, \label{eqn:symmq} \\
d_{ij} & = & ||{\mathbf y}^{(i)} - {\mathbf y}^{(j)} ||^2.\label{eqn:dist} 
\end{eqnarray}

To maximally collapsing classes, the parameters of (S-)HOPE are learned by minimizing the sum of the Kullback-Leibler divergence between the  conditional  probabilities  $q_{j|i}$ computed in the embedding space and  the ``ground-truth'' probabilities $p_{j|i}$ calculated based  on  the  class labels of training data. Specifically, $p_{j|i} \propto 1$ iff $L^{(i)} = L^{(j)}$ and $p_{j|i} = 0$ iff $L^{(i)} \neq L^{(j)}$.
Formally, the objective function of the HOPE method is as follows:
\begin{equation}\label{obj}
\small
\ell = \sum_{ij: i \neq j} p_{j|i}log\frac{p_{j|i}}{q_{j|i}} \propto -\sum_{ij: i \neq j} [L^{(i)} = L^{(j)}]logq_{j|i} + const,
\end{equation}
where $[\cdot]$ is an indicator function. 
The above objective function essentially maximizes the product of pairwise probabilities between data points in the same class, which creates favorable tight clusters that are suitable for supervised two-dimensional embedding in limited accommodable space. 
We use Conjugate Gradient Descent to optimize this objective function. Although (S-)HOPE shares the same objective as MCML~\cite{MCML2006} and dt-MCML~\cite{MinMYBZ10}, it learns a shallow explicit high-order embedding function. On the contrary, MCML aims at a linear mapping over original input features, while dt-MCML targets a complicated deep nonlinear function parametrized by a deep neural network.

\subsection{Scalable exemplar learning for data compression and fast kNN classification}
In addition to learning explicit high-order feature interactions for data embedding, we also aim to synthesize a small set of exemplars that do not exist in the training set for data compression, so that fast information retrieval such as kNN classification can be efficiently performed in the embedding space when the dataset is huge. 
Given the same dataset $\mathcal{D}$ with formal descriptions as introduced in section~\ref{sec:sup}, we aim to learn $s$ exemplars per class with their designated class labels fixed, where $s$ is a user-specified free parameter and $s \times c = z << n$. We denote these exemplars by $\{{\mathbf e}^{(j)}:  j = 1, \ldots, z\}$. We propose two approaches to exemplar learning. The first one is straightforward and relies on supervised k-means. In specific, we perform k-means on the training data to identify $s$ exemplars for each class. If a powerful feature mapping such as the one by S-HOPE is learned, all the data points in the same class will be mapped to a compact point cloud in the two-dimensional space, therefore this simple exemplar learning approach will achieve excellent performance; Otherwise, further optimization over the exemplars is needed. The second approach is based on a joint optimization. We jointly learn the high-order embedding parameters and the exemplars by optimizing the following objective function,
\begin{eqnarray}\label{exobj}
min_{{\mathbf \theta}, \{{\mathbf e}_j\}}^{} \ell({\mathbf \theta}, \{{\mathbf e}_j\}) & = & \sum_{i=1}^{n}\sum_{j=1}^{z} p_{j|i}log\frac{p_{j|i}}{q_{j|i}} 
 \propto  -\sum_{i=1}^{n} \sum_{j=1}^{z} [L^{(i)} = L^{(j)}]logq_{j|i} + const,
\end{eqnarray}
where $i$ indexes training data points, $j$ indexes exemplars, ${\mathbf \theta}$ denotes the high-order embedding parameters, $p_{j|i}$ is calculated in the same way as in section~\ref{sec:sup}, but $q_{j|i}$ is calculated with respect to exemplars,
\begin{eqnarray}
q_{j|i} & = & \frac{(1 + d_{ij})^{-1}} {\sum_{k=1}^z (1 + d_{ik})^{-1}},\\ 
d_{ij} & = & ||{\mathbf f}({\mathbf x}^{(i)}) - {\mathbf f}({\mathbf e}^{(j)}) ||^2, 
\end{eqnarray}
where ${\mathbf f}(\cdot)$ denotes the high-order embedding function as described in Equation~\ref{factensor} and \ref{shopemap}. Please note that, unlike the symmetric probability distribution in Equation~\ref{eqn:symmq},  the asymmetric $q_{j|i}$ here is computed only using the pairwise distances between training data points and exemplars. Because $z << n$, it saves us a lot of computations compared to using the original distribution in Equation~\ref{eqn:symmq}. The derivative of the above objective function with respect to exemplar ${\mathbf e}^{(j)}$ is as follows,
\begin{equation}
\frac{\partial \ell({\mathbf \theta}, {\mathbf e}_j\})} {\partial {\mathbf e}^{(j)}} = \sum_{i=1}^n 2 (1 + d_{ij})^{-1}(p_{j|i} - q_{j|i} ) ({\mathbf f}({\mathbf e}^{(j)}) - {\mathbf f}({\mathbf x}^{(i)})) \frac{\partial    {\mathbf f}({\mathbf e}^{(j)})}{\partial {\mathbf e}^{(j)}}
\end{equation}
The derivatives of other model parameters can be easily calculated similarly. We update these synthetic exemplars and the embedding parameters of HOPE in a deterministic Expectation-Maximization fashion using Conjugate Gradient Descent. 
In specific, the $s$ exemplars belonging to each class are initialized by the first exemplar learning approach. During the early phase of the joint optimization of exemplars and high-order embedding parameters, the learning process alternatively fixes one while updating the other. Then the algorithm updates all the parameters simultaneously until reaching convergence or the specified maximum number of epochs.  

\begin{table}[h]
  \centering
 \begin{tabular}{>{\quad}lc|lc}\hline
\multicolumn{2}{c|}{Linear Methods}&\multicolumn{2}{c}{Non-Linear Methods} \\ \hline
LDA&52.00&deep-AE & 24.7 \\
LPP&47.20&pt-SNE & 9.90\\
NCA&45.91&dt-NCA & 3.48\\
MCML&-&dt-MCML& 3.35\\ 
LMNN&56.28& & \\
\hline
HOPE&\textbf{5.96}&S-HOPE & \textbf{3.20} \\
\hline
\end{tabular}
  \caption{Error rates obtained by kNN (k=5) on the two-dimensional representations produced by different dimensionality reduction methods on the MNIST dataset; the result for MCML is unavailable due to its unscalability.
  }
  \label{tab:accuracy:mnist}
\end{table} 

\section{Experiments}\label{sec:experiment}
\vspace{-2mm}
In this section, we evaluate the effectiveness of (S-)HOPE by comparing it against nine different baseline methods based upon two handwritten digit datasets, i.e., MNIST and  USPS. The MNIST dataset contains 60,000 training and 10,000 test gray-level 784-dimensional images. The USPS data set contains 11,000 256-pixel gray-level images, with 8000 for training and 3000 for test. We compare the shallow linear HOPE with its five linear counterparts, including LPP, LMNN, NCA, LDA, and MCML; the non-linear shallow S-HOPE uses four deep learning baselines, including deep unsupervised models such as deep autoencoder (deep-AE)~\cite{Bengio2009} and pt-SNE~\cite{Maaten09}, as well as two deep supervised models, i.e., dt-NCA~\cite{MinMYBZ10} and dt-MCML~\cite{MinMYBZ10}. 
We set the size of exemplars as 20 in all our experiments. We used $10\%$ of training data as validation set to tune hyper-parameters such as the order $O$ of feature interactions, the number of factors ($F$), the number of high-order units ($m$), batch size, and the number of iterations for conjugate gradient descent on each mini-batch. For HOPE, $F=300$ on MNIST and $F=600$ on USPS. For S-HOPE, on MNIST, $F=400$, $m=400$; on USPS, $F=1200$, $m=400$. On both datasets, $O=3$ for HOPE and $O=2$ for S-HOPE. The parameters for all baseline methods were carefully tuned to achieve the best results. 
\subsection{Results on MNIST}
\vspace{-2mm}
\subsubsection{Classification performance on 2D embedding}
Table~\ref{tab:accuracy:mnist} presents the test error rates of 5-nearest neighbor classifier on 2-dimensional embedding generated by (S-)HOPE and the baseline methods. 
The results indicate that the linear HOPE, with an error rate of 5.96\%, significantly outperforms its linear counterparts, namely the LDA, LPP, NCA, LMNN  methods (all with an error rate about 50\%).

Promisingly, 
results in Table~\ref{tab:accuracy:mnist}  also suggest that 
our shallow method HOPE with simple Sigmoid transformations, namely S-HOPE, significantly outperforms  the  deep embedding models based on deep neural networks, in terms of accuracy obtained on the 2-dimensional embedding for visualization. For example, the error rate (3.20\%) of  S-HOPE  is lower than the ones of the deep-AE, pt-SNE, dt-NCA, and dt-MCML methods. These results clearly demonstrate the representational efficiency and power of supervised shallow models with high-order feature interactions.

To further confirm the representation power of HOPE, 
we extracted the 512-dimensional features below the softmax layer learned by a well-known deep convolutional architecture VGG~\cite{Simonyan2015}, which currently holds the-state-of-the-art classification performance through softmax layers on MNIST.  Next, we ran  S-HOPE based on these features to generate 2D embedding. Promisingly, KNN can achieve an error of 0.65\%. In contrast, NCA and LMNN on top of VGG, respectively, produces test error rate of, 1.15\% and 1.75\%. This error rate of S-HOPE  represents the historically low test error rate in two-dimensional space on MNIST. This observation implies that even with the most powerful deep learning networks, modeling explicit high-order feature interactions can achieve further predictive accuracy and outperform other models without feature interactions. 

\subsubsection{Exemplar learning}

\renewcommand{\arraystretch}{0.8}
\begin{table}[h]
  \centering
 \begin{tabular}{>{\quad}lc|lc}\hline
\multicolumn{2}{c|}{supervised kmeans}&\multicolumn{2}{c}{(S-)HOPE with exemplars } \\ \hline
 LDA&48.80\\
 LPP&45.13&supervised kmeans + HOPE& 45.29\\
 NCA&50.67&supervised kmeans + S-HOPE & \textbf{3.14}\\
 LMNN&59.67&HOPE  with 20 optimized exemplars & 5.52\\
 pt-SNE& 18.86&S-HOPE  with 20 optimized exemplars& \textbf{3.14}\\ 
 dt-MCML& 3.17 \\ 
\hline
\end{tabular}
  \caption{Error rates  obtained by 5NN on the two-dimensional representations created by different testing methods with exemplar learning on the MNIST dataset. 
  }
  \label{tab:accuracy:exemplar}
\end{table} 

In this section, we evaluate the two approaches that we propose to generating a small number of exemplars conveying high-order interactions to represent large-scale data sets. 
Table~\ref{tab:accuracy:exemplar} presents the classification errors of kNN ($k=5$) on 2-dimensional embeddings generated by (S-)HOPE with the two proposed exemplar learning.  These results suggest the following: First, using k-means as exemplar learning  works well only when coupled with S-HOPE, which demonstrates the power of the feature mapping by S-HOPE; On the other hand, when coupled with optimized exemplar learning, both HOPE and S-HOPE work very well. 
These observations suggest that, sophisticated exemplar learning method is unnecessary if we have a powerful feature mapping function such as the one by S-HOPE.

  \begin{figure}[!ht]
\subfloat[Optimized exemplars  by HOPE \label{MNIST_HOPEO3_20Exemplars}]{%
      \includegraphics[width=0.3840\textwidth]{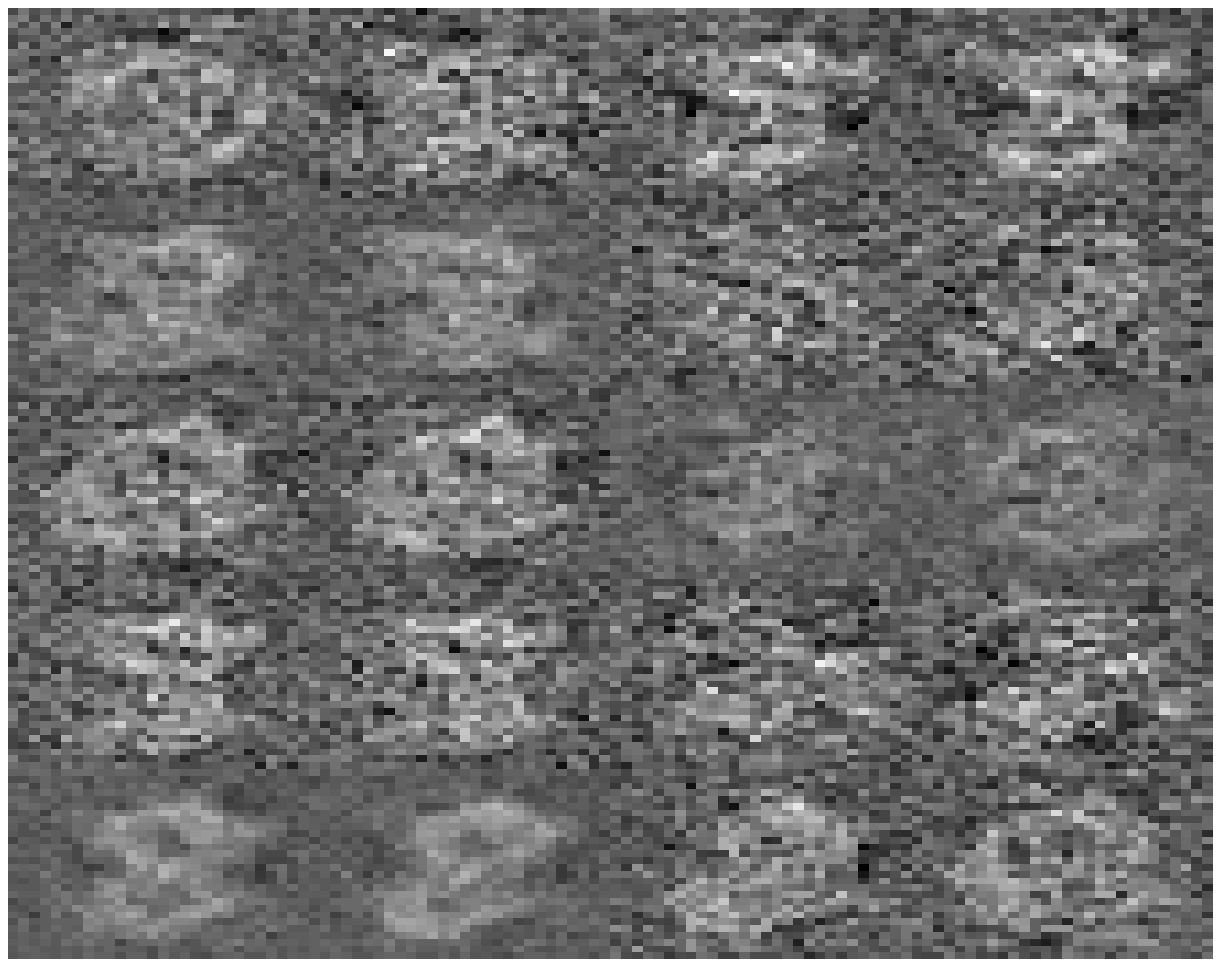}
}
        \subfloat[Optimized exemplars  by S-HOPE\label{MNIST_SHOPEO2_20Exemplars}]{%
      \includegraphics[width=0.3840\textwidth]{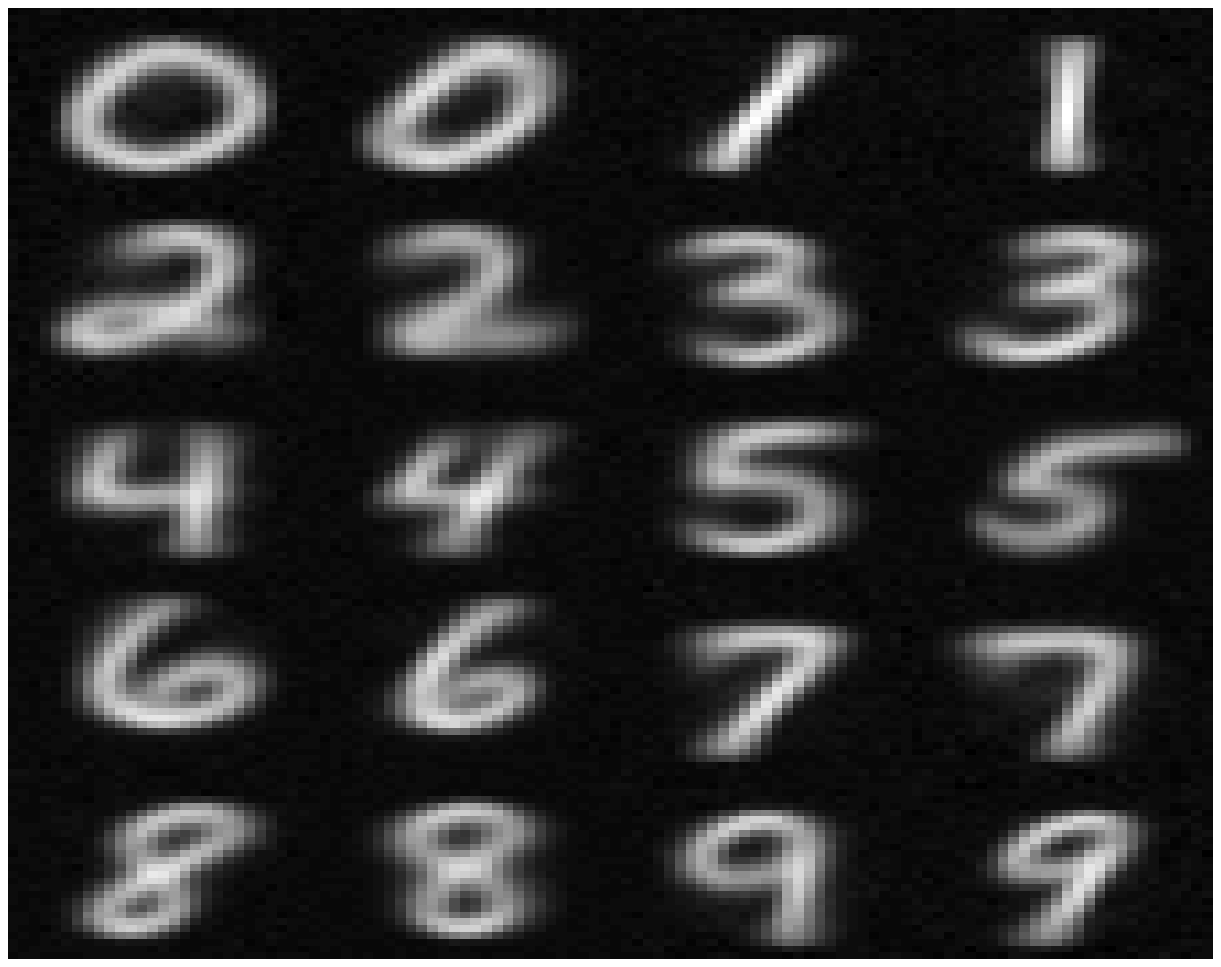}
    }
    \centering
  \caption{20 optimized exemplars created by HOPE and S-HOPE on the MNIST data set; these exemplars clearly capture essential handwritten digit variations.} 
 
  \label{fig:landmark:exemplars:mnist}
  \end{figure}

  \begin{figure*}
      \subfloat[Linear MCML\label{MNIST_MCML_embed}]{%
      \includegraphics[width=0.384\textwidth]{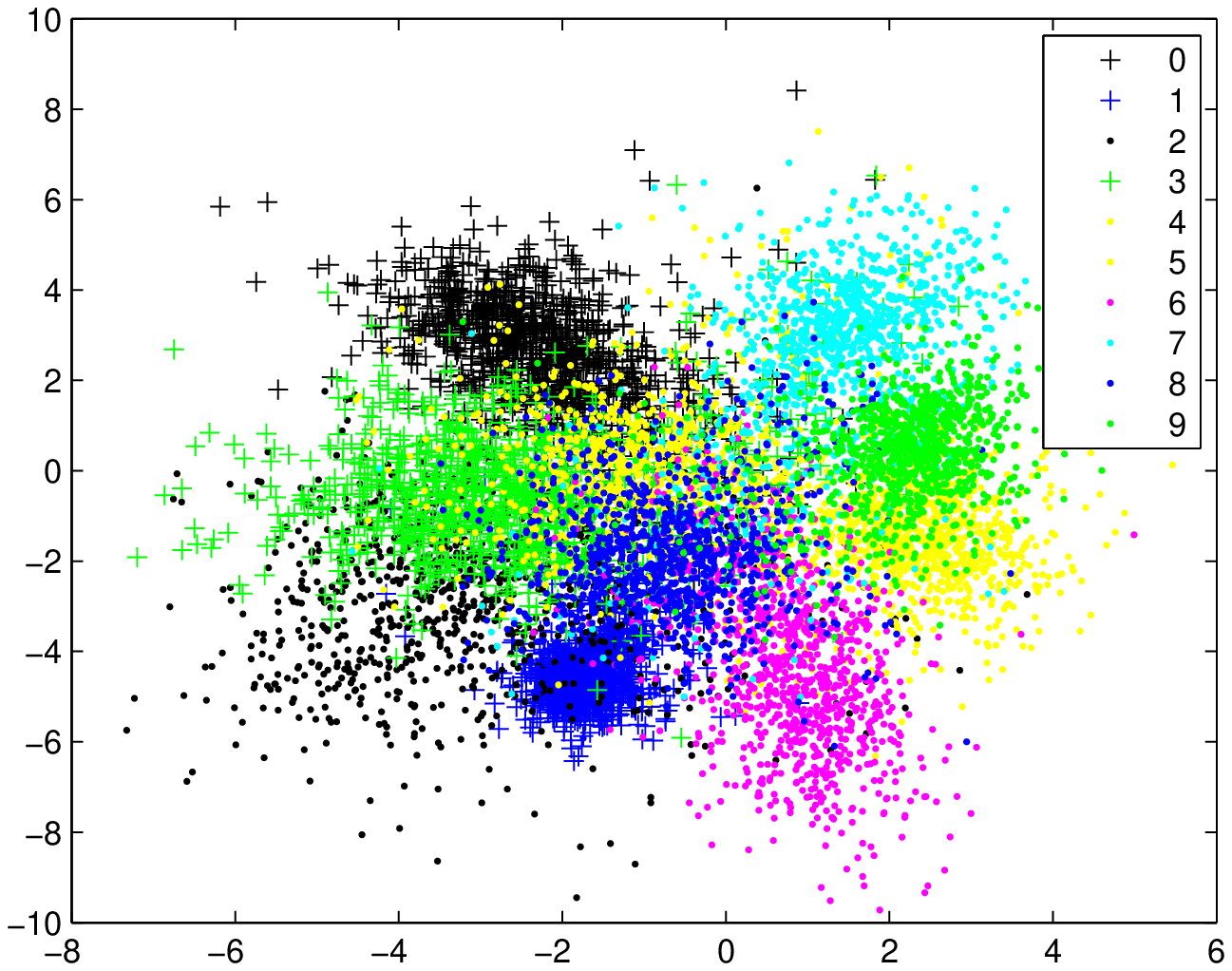}
      }    
        \subfloat[dt-MCML\label{MNIST_dtMCML_embed}]{%
      \includegraphics[width=0.384\textwidth]{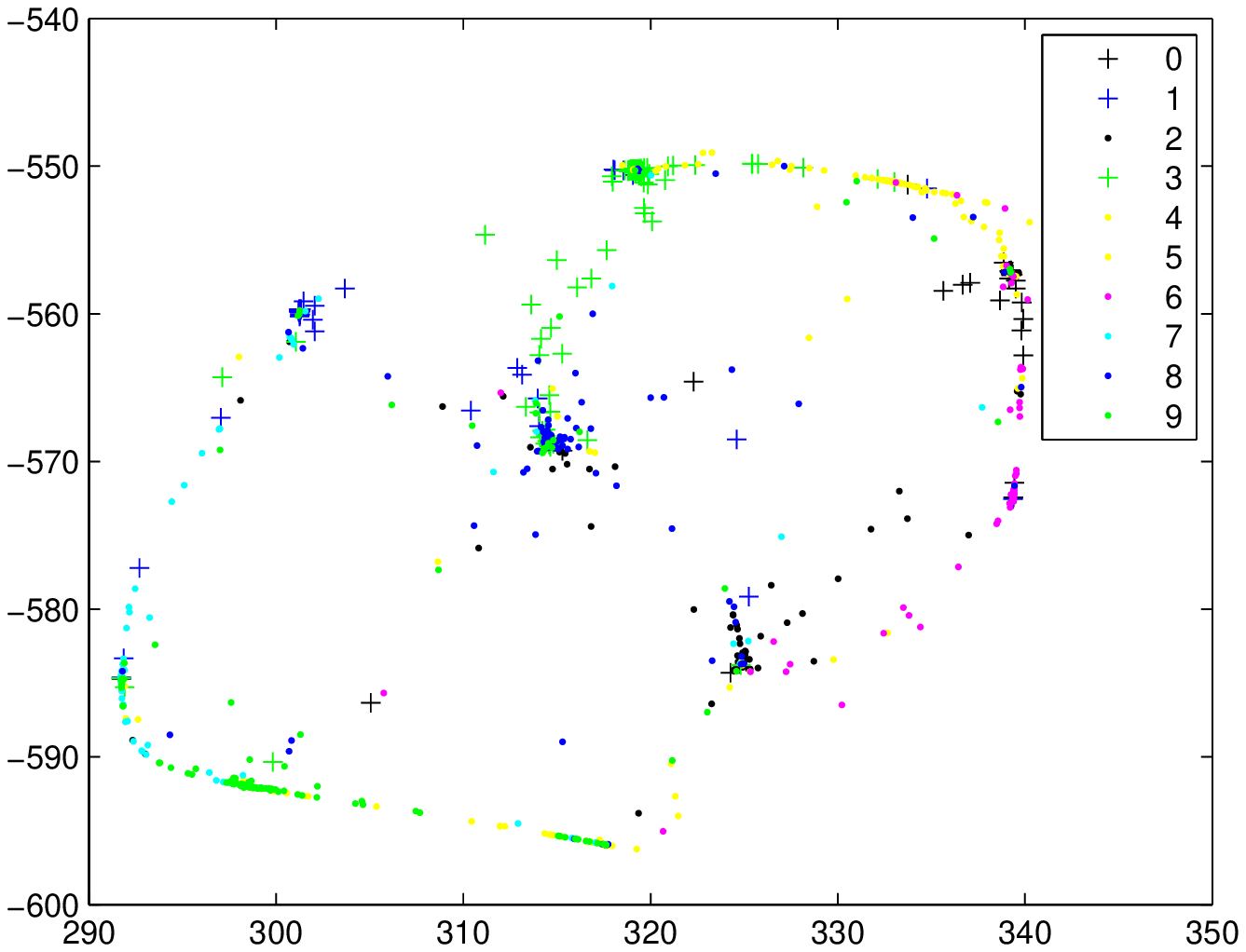}
    }
     \hfill
       \subfloat[HOPE with 20 optimized exemplars\label{MNIST_HOPEO3Exemp20_embed}]{%
      \includegraphics[width=0.384\textwidth]{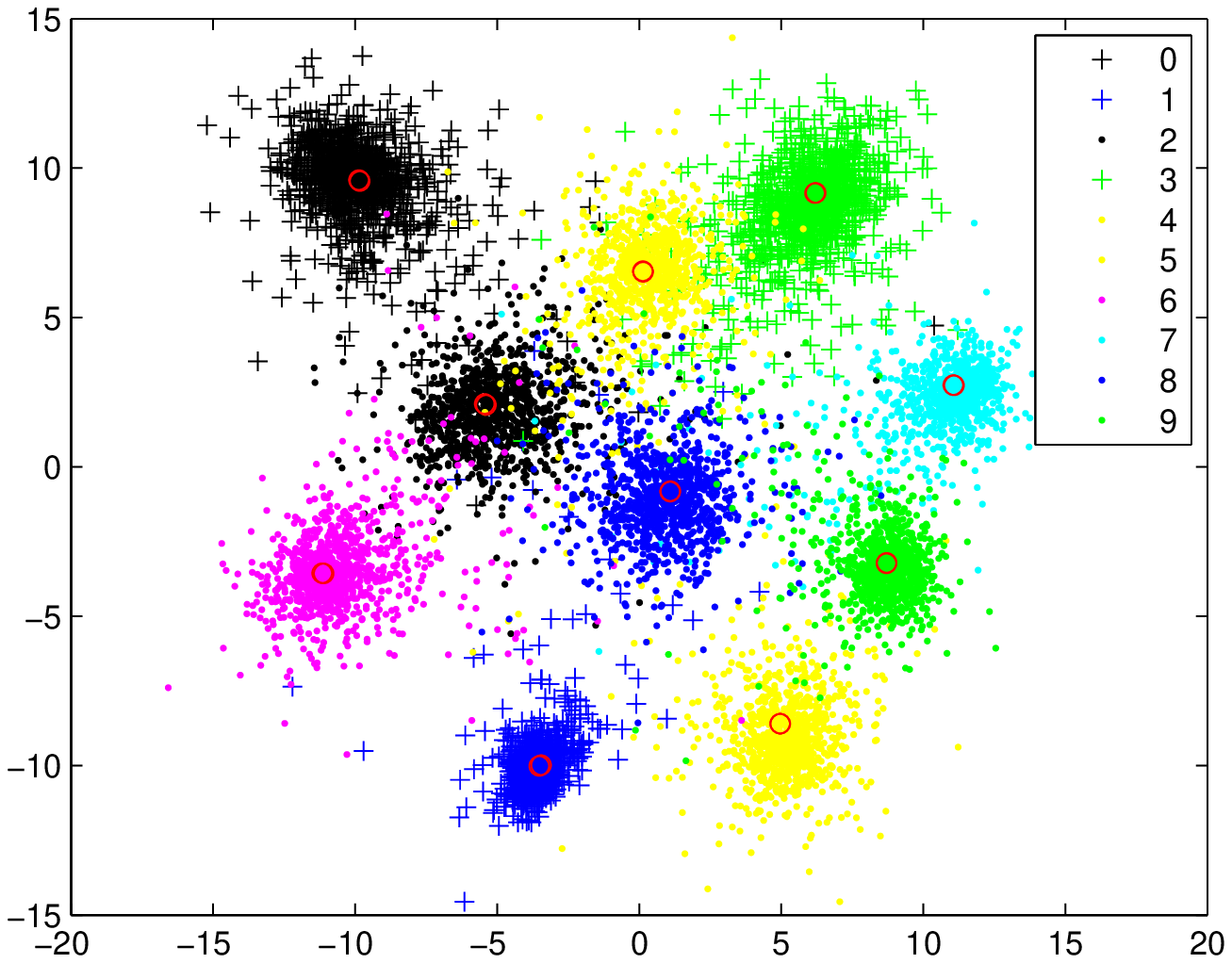}
    }   
     \subfloat[S-HOPE \label{MNIST_SHOPEO2_embed}]{%
      \includegraphics[width=0.384\textwidth]{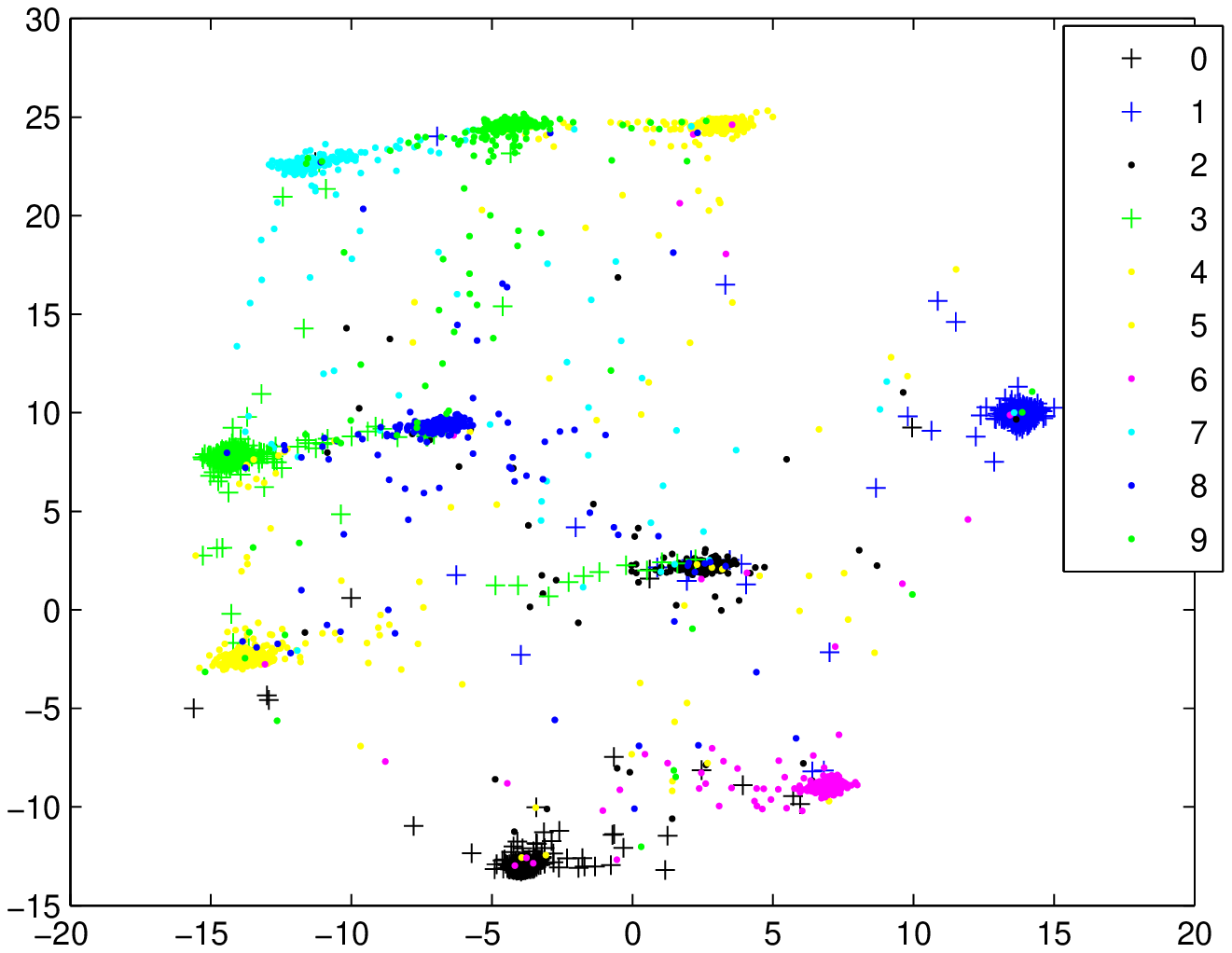}
    } 
  \centering
  \caption{Two-dimensional embeddings of 10000 MNIST test data points constructed by 
  linear MCML, dt-MCML,  HOPE-exemplar, and SHOPE; the red empty circles are the 20 optimized exemplars generated by HOPE.}
  \label{fig:visualMNIST}
  \end{figure*}


\subsubsection{Exemplars visualization} 
In Figures~\ref{fig:landmark:exemplars:mnist},  we present 20 optimized exemplars created by the most accurate models from both HOPE and S-HOPE. 
These figures indicate that S-HOPE can construct better representative exemplars than HOPE. The exemplars generated by S-HOPE clearly captured global shape information. In contrast, exemplars created by HOPE can barely be recognized by human. Part of the reason is that the former achieved much lower error (i.e., 3.14\%) than the  latter (with error of 5.52\%). Another reason is that HOPE and S-HOPE have different focus when optimizing the same cost function as depicted in Equation~\ref{exobj}. 
Promisingly, the bottom subfigure clearly show that these exemplars can capture the most important variations in the data, such as the skew and style information of different digits. Intuitively, the exemplars are based on the entire data set, thus they summarize global essential information about the data set. This is in contrast to the local knowledge contained by individual digits from a small sample when exploring massive data.

\subsubsection{2D embedding visualization}
Figures~\ref{fig:visualMNIST} shows the test data embeddings of MINST. These embeddings were constructed by, respectively, linear MCML, dt-MCML, HOPE with 20 optimized exemplars, and S-HOPE. S-HOPE produced  the  best visualization,  collapsed  all  the  data  points  in  the  same  class  close  to  each  other,  and generated large separations between class clusters. Furthermore, the embeddings of the optimized exemplars created during training (depicted as red empty circles in subfigure (c)) are  located almost at the centers of all the clusters, which suggest that the synthetic exemplars bear high-order feature interactions capturing essential data variations.

\renewcommand{\arraystretch}{0.8}
\begin{table}[h]
  \centering
 \begin{tabular}{>{\quad}lc|lc}\hline
\multicolumn{2}{c|}{Linear Methods}&\multicolumn{2}{c}{Non-Linear Methods} \\ \hline
LDA&38.23&deep-AE &28.43 \\
LPP&34.77&pt-SNE & 17.90\\
NCA&37.17&dt-NCA & 5.11\\
MCML&44.60&dt-MCML & 4.07\\ 
LMNN& 48.40&& \\
\hline
HOPE&\textbf{6.90}&S-HOPE & \textbf{3.03} \\
\hline
\end{tabular}
  \caption{Error rates  obtained by 5NN on the two-dimensional representations created by different testing methods on the USPS  dataset.
  }
  \label{tab:accuracy:usps}
\end{table} 

\renewcommand{\arraystretch}{0.8}
\begin{table}[h]
  \centering
 \begin{tabular}{>{\quad}lc|lc}\hline
\multicolumn{2}{c|}{supervised kmeans}&\multicolumn{2}{c}{(S-)HOPE with exemplars } \\ \hline
 LDA&35.23\\ 
 LPP&33.23&supervised kmeans + HOPE& 32.97\\
 NCA&35.13&supervised kmeans + S-HOPE & \textbf{2.97}\\
 LMNN&59.67&HOPE  with 20 optimized exemplars & 6.90\\
 pt-SNE& 29.47&S-HOPE  with 20 optimized exemplars& 3.60\\ 
 dt-MCML& 4.27\\
\hline
\end{tabular}
  \caption{Error rates  obtained by 5NN on the two-dimensional representations created by different testing methods with exemplar learning on the USPS dataset. 
  }
  \label{tab:accuracy:exemplar:usps}
\end{table}

\subsection{Results on USPS}
We also conducted experiments on the USPS data set. Table~\ref{tab:accuracy:mnist} presents the performances of kNN classification ($k = 5$) on two-dimensional embeddings constructed by various dimensionality reduction techniques. 

From results as presented in Tables~\ref{tab:accuracy:usps} and~\ref{tab:accuracy:exemplar:usps}, one can draw very similar conclusions as the ones on the MNIST data. 
Visualization on the exemplars and embeddings learned also show consistent behaviors of the (S-)HOPE models as that on the MNIST data. We included all the plotted images with high-resolution in the supplementary material to this paper.

\subsection{Computational efficiency of exemplar learning}
S-HOPE with exemplar learning speeds up computational efficiency for fast information retrieval such as kNN classification used in the above experiments by thousands of times. On MNIST and USPS, in the feature space, test data prediction is against, respectively, 60000 training data points in 784-dimensional space (test error rate 3.05\%) and 8000 training data points in 256-dimensional space (test error rate 4.77\%). With S-HOPE and 20 synthesized exemplars, test data prediction is only against 20 exemplars in 2-dimensional space and even gets comparable or much better performance than in the original feature space! This computational speedup will be more pronounced on massive datasets. 
\section{Conclusion and future work}
\label{sec:discussion}
In this paper, we present a supervised High-Order Parametric Embedding (HOPE) approach for data visualization and compression. Our experimental results indicate that modeling high-order feature interactions  
can significantly improve the data visualization in low-dimensional embedding space, when compared with its linear counterparts.  
Surprisingly, our shallow method HOPE with simple Sigmoid transformations significantly outperforms state-of-the-art supervised deep embedding models based on deep neural networks, and even achieved historically low test error rate of 0.65\% in two-dimensional space on MNIST. In addition, the learned synthetic exemplars in combination with the shallow high-order feature mapping speed up kNN classification by thousands of times with comparable or much better performance than that in the original feature space. These results clearly demonstrate the high representational efficiency and power of supervised shallow models with high-order feature interactions, and suggest that the performance and representational efficiency of supervised deep learning models might be significantly improved by incorporating explicit high-order feature interactions. 
Our methods can be readily extended to the setting of unsupervised learning, for which we just need to compute the pairwise probabilities  $p_{j|i}$'s using high-dimensional feature vectors instead of class labels and optimize exemplars accordingly. S-HOPE can also be easily extended to deep structures across different layers.

\begin{spacing}{0.6}
\setlength{\bibsep}{3pt}
\bibliography{ref}
\bibliographystyle{abbrv}
\end{spacing}
\end{document}